\documentclass[11pt,a4paper]{article}
\usepackage[hyperref]{acl2018}
\usepackage{times}
\usepackage{latexsym}
\usepackage{pbox}
\usepackage{times}
\usepackage{url}
\usepackage{latexsym}
\usepackage[utf8]{inputenc}
\usepackage{graphicx}
\usepackage{fixltx2e}
\usepackage{textcomp}
\usepackage{array}
\usepackage{color}
\usepackage{amsmath}
\usepackage{multirow}
\usepackage{floatrow}
\usepackage{float}
\usepackage{soul}
\usepackage{natbib}
\usepackage{longtable}
\usepackage{listingsutf8}
\usepackage{color, colortbl}
\usepackage{empheq}
\usepackage{makecell}

\usepackage{subcaption}
\usepackage{etoolbox}
\usepackage{pgfplots}
\usepackage{color}
\definecolor{higher}{rgb}{0.0, 0.5, 0.0}
\definecolor{lower}{rgb}{1.0, 0.01, 0.24}

\usepackage{url}

\aclfinalcopy % Uncomment this line for the final submission
\def\aclpaperid{964} %  Enter the acl Paper ID here

\title{Character-Level Models versus Morphology in Semantic Role Labeling}

\author{Gözde Gül Şahin \\
  Department of Computer Science \\
  Technische Universität Darmstadt \\
  Darmstadt, Germany \\
  {\tt isguderg@itu.edu.tr} \\\And
  Mark Steedman \\
  School of Informatics \\
  University of Edinburgh  \\
  Edinburgh, Scotland \\
  {\tt steedman@inf.ed.ac.uk} \\}

\date{}

\begin{document}
\maketitle
\begin{abstract}

Character-level models have become a popular approach specially for their accessibility and ability to handle unseen data. However, little is known on their ability to reveal the underlying morphological structure of a word, which is a crucial skill for high-level semantic analysis tasks, such as semantic role labeling (SRL). In this work, we train various types of SRL models that use word, character and morphology level information and analyze how performance of characters compare to words and morphology for several languages. We conduct an in-depth error analysis for each morphological typology and analyze the strengths and limitations of character-level models that relate to out-of-domain data, training data size, long range dependencies and model complexity. Our exhaustive analyses shed light on important characteristics of character-level models and their semantic capability. 
\end{abstract}

\section{Introduction}

  % Encoding of words is important, word embeddings have issues
  % Issues are: OOV and unable to catch regularities
  Encoding of words is perhaps the most important step towards a successful end-to-end natural language processing application. Although word embeddings have been shown to provide benefit to such models, they commonly treat words as the smallest meaning bearing unit and assume that each word type has its own vector representation. This assumption has two major shortcomings especially for languages with rich morphology:~(1)~inability to handle unseen or out-of-vocabulary (OOV) word-forms (2) inability to exploit the regularities among word parts. %For instance the commonalities between infrequent words \textit{\textbf{vergi}lendirebildiklerinden} and \textit{\textbf{belge}lendirebildiklerinden (from the ones that they could document)} will be very hard to capture.
 
  % Issues become more important on semantic tasks
  % Word part - semantic çok ilişkili
  % Dolayısıyla word partların semantic olarak daha başarılı olacağını düşünüyoruz
  The limitations of word embeddings are particularly pronounced in sentence-level semantic tasks, especially in languages where word parts play a crucial role. Consider the Turkish sentences ``\textit{Köy+\textbf{lü-ler} (villagers) şehr+\textbf{e} (to town) geldi (came)}'' and ``\textit{Sendika+\textbf{lı-lar} (union members) meclis+\textbf{e} (to council) geldi (came)}''. Here the stems \textit{köy (village)} and \textit{sendika (union)} function similarly in semantic terms with respect to the verb \textit{come} (as \textit{the origin of the agents of the verb}), where \textit{şehir (town)} and \textit{meclis (council)} both function as \textit{the end point}. These semantic similarities are determined by the common word parts shown in \textbf{bold}. However ortographic similarity does not always correspond to semantic similarity. For instance the ortographically similar words \textit{knight} and \textit{night} have large semantic differences. Therefore, for a successful semantic application, the model should be able to capture both the regularities, \textit{i.e, morphological tags} and the irregularities, \textit{i.e, lemmas} of the word. 
  
  Morphological analysis already provides the aforementioned information about the words. However access to useful morphological features may be problematic due to software licensing issues, lack of robust morphological analyzers and high ambiguity among analyses. Character-level models (CLM), being a cheaper and accessible alternative to morphology, have been reported as performing competitively on various NLP tasks~\cite{ling2015finding,plank2016multilingual,lee2016fully}. However the extent to which these tasks depend on morphology is small; and their relation to semantics is weak. Hence, little is known on their true ability to reveal the underlying morphological structure of a word and their semantic capabilities. Furthermore, their behaviour across languages from different families; and their limitations and strengths such as handling of long-range dependencies, reaction to model complexity or performance on out-of-domain data are unknown. Analyzing such issues is a key to fully understanding the character-level models. 

  To achieve this, we perform a case study on semantic role labeling (SRL), a sentence-level semantic analysis task that aims to identify predicate-argument structures and assign meaningful labels to them as follows: 
  \begin{quote} 
  $[$Villagers$]$\textsubscript{comers} came $[$to town$]$\textsubscript{end point}
  \end{quote} 
  We use a simple method based on bidirectional LSTMs to train three types of base semantic role labelers that employ (1) words (2) characters and character sequences and (3) gold morphological analysis. The gold morphology serves as the upper bound for us to compare and analyze the performances of character-level models on languages of varying morphological typologies. We carry out an exhaustive error analysis for each language type and analyze the strengths and limitations of character-level models compared to morphology. In regard to the diversity hypothesis which states that \textit{diversity} of systems in ensembles lead to further improvement, we combine character and morphology-level models and measure the performance of the ensemble to better understand how similar they are. 

  We experiment with several languages with varying degrees of morphological richness and typology: Turkish, Finnish, Czech, German, Spanish, Catalan and English. Our experiments and analysis reveal insights such as:
  \begin{itemize}
  \item CLMs provide great improvements over whole-word-level models despite not being able to match the performance of morphology-level models (MLMs) for \textit{in-domain} datasets. However their performance surpass all MLMs on \textit{out-of-domain} data,
  \item Limitations and strengths differ by morphological typology. Their limitations for agglutinative languages are related to rich \textit{derivational morphology} and high \textit{contextual ambiguity}; whereas for fusional languages they are related to \textit{number of morphological tags} (morpheme ambiguity) ,    
  %\item Similarity between character and morphology-level models is higher than the similarity within character-level (char and char-trigram) models on languages with high OOV\% and vice versa,
  \item CLMs can handle long-range dependencies equally well as MLMs,
  \item In presence of more training data, CLM's performance is expected to improve faster than of MLM.
  %\item Morphology-level models benefit more from increasing model complexity.
  \end{itemize}

  % However, it is manifestly clear that similarity in form is neither a necessary nor sufficient condition for similarity in function: small orthographic differences may correspond to large semantic or syntactic differences (butter vs. batter), and large orthographic differences may obscure nearly perfect functional correspondence (rich vs. affluent). Thus, any orthographically aware model must be able to capture non-compositional effects in addition to more regular effects due to, e.g., morphological processes. To model the complex form–function relationship, we turn to long short-term memories (LSTMs), which are designed to be able to capture complex non-linear and non-local dynamics in sequences.

\section{Related Work}
  \paragraph{Neural SRL Methods:}
  Neural networks have been first introduced to the SRL scene by \citet{Collobert2011}, where they use a unified end-to-end convolutional network to perform various NLP tasks. Later, the combination of neural networks (LSTMs in particular) with traditional SRL features (categorical and binary) has been introduced~\cite{fitzgerald2015semantic}. Recently, it has been shown that careful design and tuning of deep models can achieve state-of-the-art with no or minimal syntactic knowledge for English and Chinese SRL. Although the architectures vary slightly, they are mostly based on a variation of bi-LSTMs. \citet{zhou2015end,he2017deep} connect the layers of LSTM in an interleaving pattern where in~\cite{wang2015chinese,diegosimple} regular bi-LSTM layers are used. Commonly used features for the encoding layer are: pretrained word embeddings; distance from the predicate; predicate context; predicate region mark or flag; POS tag; and predicate lemma embedding. Only a few of the models~\cite{diegosimple,diegograph} perform dependency-based SRL. Furthermore, all methods focus on languages with rich resources and less morphological complexity like English and Chinese.
  \paragraph{Character-level Models:}
   Character-level models have proven themselves useful for many NLP tasks such as language modeling~\cite{ling2015finding,kim2016character}, POS tagging~\cite{santos2014learning,plank2016multilingual}, dependency parsing~\cite{dozat2017stanford} and machine translation~\cite{lee2016fully}. However the number of comparative studies that analyze their relation to morphology are rather limited. Recently, \citet{vania2017characters} presented a unified framework, where they investigated the performances of different subword units, namely characters, morphemes and morphological analysis on language modeling task. They experimented with languages of varying morphological typologies and concluded that the performance of character models can not yet match the morphological models, albeit very close. Similarly,~\citet{belinkov2017neural} analyzed how different word representations help learn better morphology and model rare words on a neural MT task and concluded that character-based representations are much better for learning morphology.

\section{Method} 
 Formally, we generate a label sequence $\vec{l}$ for each sentence and predicate pair: $(s,p)$. Each $l_t\in\vec{l}$ is chosen from $\mathcal{L}=\{ \mathit{roles \cup nonrole}\}$, where $roles$ are language-specific semantic roles (mostly consistent with PropBank) and $nonrole$ is a symbol to present tokens that are not arguments. Given $\theta$ as model parameters and $g_t$ as gold label for $t_{th}$ token, we find the parameters that minimize the negative log likelihood of the sequence:
  \begin{gather} 
    \hat{\theta}=\underset{\theta}{\arg\min} \left( -\sum_{t=1}^n log (p(g_t|\theta,s,p)) \right)
  \end{gather}
  Label probabilities, $p(l_t|\theta,s,p)$, are calculated with equations given below.%the proposed model shown in Fig.~\ref{fig:LSTM_model}.
  % \begin{figure*}
  %  \centering  
  %  \includegraphics[scale=0.55]{figures/lstm_paper.pdf}
  %  \caption{Labeling of semantic roles for input pair $s=$``Uzanıp mektubu aldı'' \textit{(He/she reached out and grabbed the letter.)} and $p=$``al'' \textit{(to grab)}. $\phi$ refers to a bi-LSTM unit, pf is predicate flag.}
  %  \label{fig:LSTM_model}
  % \end{figure*}
  First, the word encoding layer splits tokens into subwords via $\rho$ function.
  \begin{gather} 
    \rho(w) = {s_0,s_1,..,s_n}
  \end{gather}
  As proposed by \citet{ling2015finding}, we treat words as a sequence of subword units. Then, the sequence is fed to a simple bi-LSTM network~\cite{graves2005framewise,lstmpaper} and hidden states from each direction are weighted with a set of parameters which are also learned during training. Finally, the weighted vector is used as the word embedding given in Eq.~\ref{eq:comp}.
  \begin{gather} 
    hs_f, hs_b = \text{bi-LSTM}({s_0,s_1,..,s_n}) \\
    \vec{w} = W_f \cdot hs_f + W_b \cdot hs_b + b
  \label{eq:comp}
  \end{gather}
  There may be more than one predicate in the sentence so it is crucial to inform the network of which arguments we aim to label. In order to mark the predicate of interest, we concatenate a predicate flag $pf_t$ to the word embedding vector. 
  \begin{gather} 
    \vec{x_{t}} = [\vec{w};pf_t] 
  \end{gather}
  Final vector, $\vec{x_t}$ serves as an input to another bi-LSTM unit.
  \begin{gather} 
    \vec{h_{f}, h_{b}} = \text{bi-LSTM}(x_{t})
  \end{gather}
  Finally, the label distribution is calculated via softmax function over the concatenated hidden states from both directions.
  \begin{gather} 
    \vec{p(l_t|s,p)} = softmax(W_{l}\cdot[\vec{h_{f}};\vec{h_{b}}]+\vec{b_{l}})
  \end{gather}
  For simplicity, we assign the label with the highest probability to the input token.~\footnote{Our implementation can be found at \url{https://github.com/gozdesahin/Subword_Semantic_Role_Labeling}}. 

 \subsection{Subword Units}
 \label{ssec:units}
   We use three types of units: (1) words (2) characters and character sequences and (3) outputs of morphological analysis. Words serve as a lower bound; while morphology is used as an upper bound for comparison. Table~\ref{tab:subunits} shows sample outputs of various $\rho$ functions. 
   \begin{table}[!ht]
     \scalebox{0.75}
     {
     \begin{tabular}{llll}     
      \textbf{$\rho$} & \textbf{word} & \textbf{output} \\
          \hline
          \small \textit{char} & available & $<$-a-v-a-i-l-a-b-l-e-$>$ \\
          \small \textit{char3} & available & $<$av-ava-vai-ail-ila-lab-abl-ble-le$>$\\
          \hline
          \small \textit{morph-DEU} & prächtiger & [\textit{prächtig};Pos;Nom;Sg;Masc] \\
          \small \textit{morph-SPA} & las & [\textit{el};postype=article;gen=f;num=p] \\
          \small \textit{morph-CAT} & la & [\textit{el};postype=article;gen=f;num=s] \\
          \small \textit{morph-TUR} & boyundaki & [\textit{boy};\textsc{Noun};A3sg;P3sg;Loc;DB;\textsc{Adj}] \\
          \small \textit{morph-FIN} & tyhjyyttä & [\textit{tyhjyys};Case=Par;Number=Sing] \\
          \small \textit{morph-CZE} & si & [\textit{se};SubPOS=7;Num=X;Cas=3] \\
     \end{tabular}
     }
   \caption{Sample outputs of different \textbf{$\rho$} functions} 
   \label{tab:subunits}
   \end{table}
   Here, \textit{char} function simply splits the token into its characters. Similar to n-gram language models, \textit{char3} slides a character window of width $n=3$ over the token. Finally, gold morphological features are used as outputs of \textit{morph-language}. Throughout this paper, we use \textit{morph} and \textit{oracle} interchangably, i.e., morphology-level models (MLM) have access to gold tags unless otherwise is stated. For all languages, \textit{morph} outputs the \textit{lemma} of the token followed by language specific morphological tags. As an exception, it outputs additional information for some languages, such as parts-of-speech tags for Turkish. Word segmenters such as Morfessor and Byte Pair Encoding (BPE) are other commonly used subword units. Due to low scores obtained from our preliminary experiments and unsatisfactory results from previous studies~\cite{vania2017characters}, we excluded these units. 

\section{Experiments}
\label{ch5sec:exper}
    We use the datasets distributed by LDC for Catalan (CAT), Spanish (SPA), German (DEU), Czech (CZE) and English (ENG)~\cite{LDC2012T03,LDC2012T04}; and datasets made available by~\citet{Haverinen2015,csahinannotation} for Finnish (FIN) and Turkish (TUR) respectively~\footnote{Turkish PropBank is based on previous efforts~\cite{td1,td2,td3,td4,td5,td6}}. Datasets are provided with syntactic dependency annotations and semantic roles of verbal predicates. In addition, English supplies nominal predicates annotated with semantic roles and does not provide any morphological feature.
    \begin{table}
        \scalebox{0.85}
        {
          \begin{tabular}{lrrrrccc}  
           &  \textbf{\#sent} & \textbf{\#token} & \textbf{\#pred} & \textbf{\#role}  & \textbf{type} \\
          \hline 
          \textbf{CZE} & 39K & 653K & 414K & 51 & \textsc{F} \\ 
          \textbf{ENG} & 39K & 958K & 179K & 38 & \textsc{F} \\ 
          \textbf{DEU} & 36K & 649K & 17K & 9 & \textsc{F} \\ 
          \textbf{SPA} & 14K & 419K & 44K & 34 & \textsc{F} \\          
          \textbf{CAT} & 13K & 384K & 37K & 35 & \textsc{F} \\ 
          \textbf{FIN} & 12K & 163K & 27K & 20 & \textsc{A} \\ 
          \textbf{TUR} & 4K & 39K & 8K & 26 & \textsc{A} \\ 
          \end{tabular}
         }
      \caption{Training data statistics. \textsc{A}: Agglutinative, \textsc{F}: Fusional}
      \label{tab:stats}
    \end{table} 
    Statistics for the training split for all languages are given in Table~\ref{tab:stats}. Here, \textbf{\#pred} is number of predicates, and \textbf{\#role} refers to number distinct semantic roles that occur more than 10 times. More detailed statistics about the datasets can be found in~\citet{conll09,Haverinen2015,csahinannotation}.  
    
    \subsection{Experimental Setup}
      %For some languages predicate senses are not provided, and for some, around 81\% of the predicates are annotated with first sense. Therefore we focus on identifying and labeling arguments.
      To fit the requirements of the SRL task and of our model, we performed the following:
      \paragraph{Spanish, Catalan:} Multiword expressions (MWE) are represented as a single token, \textit{(e.g., Confederación\_Francesa\_del\_Trabajo)}, that causes notably long character sequences which are hard to handle by LSTMs. For the sake of memory efficiency and performance, we used an abbreviation \textit{(e.g., CFdT)} for each MWE during training and testing.  
      \paragraph{Finnish:} Original dataset defines its own format of semantic annotation, such as 17:PBArgM\_mod$\mid$19:PBArgM\_mod meaning the node is an argument of $17_{th}$ and $19_{th}$ tokens with \textit{ArgM-mod} (temporary modifier) semantic role. They have been converted into CoNLL-09 tabular format, where each predicate's arguments are given in a specific column. 
      \paragraph{Turkish:} Words are splitted from derivational boundaries in the original dataset, where each inflectional group is represented as a separate token. We first merge boundaries of the same word, \textit{i.e, tokens of the word}, then we use our own $\rho$ function to split words into subwords. 
      \paragraph{Training and Evaluation:} We lowercase all tokens beforehand and place special start and end of the token characters. For all experiments, we initialized weight parameters orthogonally and used one layer bi-LSTMs both for subword composition and argument labeling with hidden size of 200. Subword embedding size is chosen as 200. We used gradient clipping and early stopping to prevent overfitting. Stochastic gradient descent is used as the optimizer. The initial learning rate is set to 1 and reduced by half if scores on development set do not improve after 3 epochs. We use the provided splits and evaluate the results with the official evaluation script provided by CoNLL-09 shared task. In this work (and in most of the recent SRL works), only the scores for argument labeling are reported, which may cause confusions for the readers while comparing with older SRL studies. Most of the early SRL work report combined scores (argument labeling with predicate sense disambiguation (PSD)). However, PSD is considered a simpler task with higher F1 scores~\footnote{For instance in English CoNLL-09 dataset, 87\% of the predicates are annotated with their first sense, hence even a dummy classifier would achieve 87\% accuracy. The best system from CoNLL-09 shared task reports 85.63 F1 on English evaluation dataset, however when the results of PSD are discarded, it drops down to 81.}. Therefore, we believe omitting PSD helps us gain more useful insights on character level models.  

     \begin{figure}
      \centering
      \begin{subfigure}[b]{0.95\textwidth}
          \includegraphics[width=\textwidth]{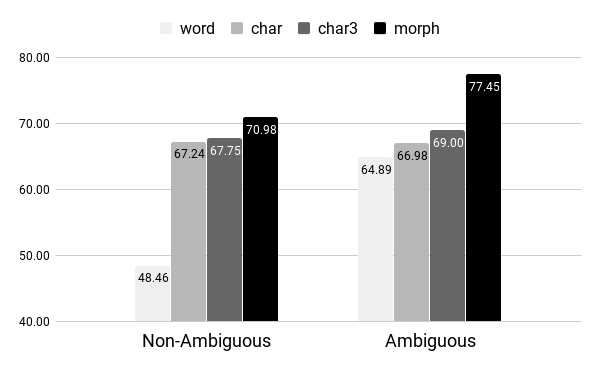}
          \caption{Finnish - Contextual ambiguity}
          \label{fig:fin}
      \end{subfigure}
      ~ %add desired spacing between images, e. g. ~, \quad, \qquad, \hfill etc. 
      \begin{subfigure}[b]{0.95\textwidth}
          \includegraphics[width=\textwidth]{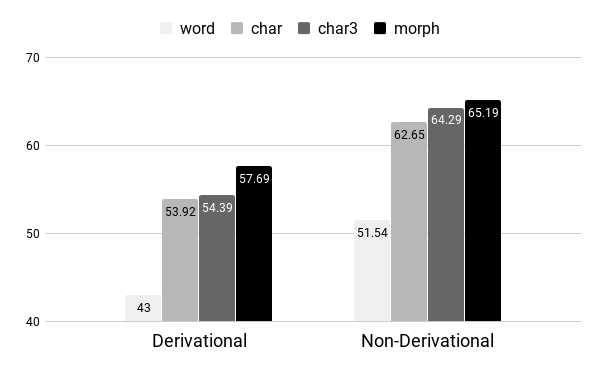}
          \caption{Turkish - Derivational morphology}
          \label{fig:tur}
      \end{subfigure}
      \caption{Differences in model performances on agglutinative languages}
    \label{fig:agg_issue}
    \end{figure} 

\section{Results and Analysis}
  Our main results on test and development sets for models that use words, characters (\textit{char}), character trigrams (\textit{char3}) and morphological analyses (\textit{morph}) are given in Table~\ref{tab:arglabel_res}. We calculate \textit{improvement over word (IOW)} for each subword model and \textit{improvement over the best character model (IOC)} for the \textit{morph}. \textit{IOW} and \textit{IOC} values are calculated on the test set.    
  \begin{table}
    \scalebox{0.65}
    {
      \begin{tabular}{|r|c|c|c|c|c|c|c|c|}
       \hline
       & \textbf{word} & \multicolumn{2}{c|}{\textbf{char}} & \multicolumn{2}{c|}{\textbf{char3}} & \multicolumn{3}{c|}{\textbf{morph}}\\
       \hline
         & F1 & F1 & \textit{\small IOW\%} & F1 & \textit{ \small IOW\%} & F1 & \textit{\small IOW\%} & \textit{\small IOC\%} \\
         \hline 
         \multirow{2}{*}{\rotatebox[origin=c]{90}{\textbf{FIN}}} & 48.91 & 67.24 & \multirow{2}{*}{\textit{37.46}} & 67.78 & \multirow{2}{*}{\textit{38.58}} & \textbf{71.15} & \multirow{2}{*}{\textit{45.47}} & \multirow{2}{*}{\textit{4.97}} \\ 
                                       & 51.65 & 66.82 & & 67.08 &  & \textbf{71.88} &  & \\ 

         \hline
         \multirow{2}{*}{\rotatebox[origin=c]{90}{\textbf{TUR}}} & 44.82 & 55.89 & \multirow{2}{*}{\textit{24.68}} & 56.60 & \multirow{2}{*}{\textit{26.28}} & \textbf{59.38} & \multirow{2}{*}{\textit{32.48}} & \multirow{2}{*}{\textit{4.91}}\\  
                                       & 43.14 & 54.48 & & 55.41 & & \textbf{58.91} &  & \\ 

         \hline
         \multirow{2}{*}{\rotatebox[origin=c]{90}{\textbf{SPA}}} & 64.30 & 67.90 & \multirow{2}{*}{\textit{5.61}} & 68.43 & \multirow{2}{*}{\textit{6.42}} & \textbf{69.39} & \multirow{2}{*}{\textit{7.92}} & \multirow{2}{*}{\textit{2.25}}\\
                                       & 64.53 & 67.64 & & 67.64 & & \textbf{69.17} & & \\
         \hline

         \multirow{2}{*}{\rotatebox[origin=c]{90}{\textbf{CAT}}} & 65.45 & 70.56 & \multirow{2}{*}{\textit{7.82}} & 71.34 & \multirow{2}{*}{\textit{9.00}} & \textbf{73.24} & \multirow{2}{*}{\textit{11.90}} & \multirow{2}{*}{\textit{2.66}} \\ 
                                       & 65.67 & 70.43 & & 70.48 &  & \textbf{72.36} & & \\ 

         \hline
         \multirow{2}{*}{\rotatebox[origin=c]{90}{\textbf{CZE}}} & 63.58 & 74.04 & \multirow{2}{*}{\textit{16.45}} & 74.98 & \multirow{2}{*}{\textit{17.93}} & \textbf{80.66} & \multirow{2}{*}{\textit{26.87}} & \multirow{2}{*}{\textit{7.58}}\\    
                                       & 72.69 & 74.58 &  & 75.59 & & \textbf{81.06} &  & \\ 
        
         \hline             
         \multirow{2}{*}{\rotatebox[origin=c]{90}{\textbf{DEU}}} & 54.78 & 63.71 & \multirow{2}{*}{\textit{16.29}} & 65.56 & \multirow{2}{*}{\textit{19.68}} & \textbf{69.35} & \multirow{2}{*}{\textit{26.58}} & \multirow{2}{*}{\textit{5.77}}\\      
                                       & 53.76 & 62.75 & & 63.70 &  & \textbf{72.18} & & \\ 

         \hline
         \multirow{2}{*}{\rotatebox[origin=c]{90}{\textbf{ENG}}} & 81.19 & \textbf{81.61} & \multirow{2}{*}{\textit{0.52}} & 80.65 & \multirow{2}{*}{\textit{-0.67}} & - & - & - \\  
                                       & 78.67 & \textbf{79.22} &  & 78.85 & & - & - & - \\  
         \hline
       \end{tabular}
    }
    \caption{F1 scores of word, character, character trigram and morphology models for argument labeling. Best F1 for each language is shown in \textbf{bold}. First row: results on test, Second row: results on development.}
  \label{tab:arglabel_res}
  \end{table}

  The biggest improvement over the word baseline is achieved by the models that have access to morphology for all languages~(except for English) as expected. Character trigrams consistently outperformed characters by a small margin. Same pattern is observed on the results of the development set. \textit{IOW} has the values between 0\% to 38\% while \textit{IOC} values range between 2\%-10\% dependending on the properties of the language and the dataset. We analyze the results separately for agglutinative and fusional languages and reveal the links between certain linguistic phenomena and the \textit{IOC}, \textit{IOW} values.
  \paragraph{Agglutinative languages} have many morphemes attached to a word like beads on a string. This leads to high number of OOV words and cause word lookup models to fail. Hence, the highest \textit{IOW}s by character models are achieved on these languages: Finnish and Turkish. This language family has one-to-one morpheme to meaning mapping with small orthographic differences~\textit{(e.g., mış, miş, muş, müş for past perfect tense)}, that can be easily extracted from the data. Even though each morpheme has only one interpretation, each word (consisting of many morphemes) has usually more than one. For instance two possible analyses for the Turkish word ``dolar'' are (1) ``dol+Verb+Positive+Aorist+3sg''~\textit{(it fills)}, (2) ``dola+Verb+Positive+Aorist+3sg''~\textit{(he/she wraps)}. For a syntactic task, models are not obliged to learn the difference between the two; whereas for a semantic task like SRL, they are. We will refer to this issue as \textit{contextual ambiguity}. Another important linguistic issue for agglutinative languages is the complex interaction between morphology and syntax, which is usually achieved via derivational morphemes. In other words, unlike \textit{inflectional} morphemes that only give information on \textit{word-level semantics}, derivational morphemes provide more clues on \textit{sentence-level semantics}. The effects of these two phenomena on model performances is shown in Fig.~\ref{fig:agg_issue}. Scores given in Fig.~\ref{fig:agg_issue} are absolute F1 scores for each model. For the analysis in Fig.~\ref{fig:fin}, we separately calculated F1 scores of each model on words that have been observed with at least two different set of morphological features (\textit{ambiguous}), and one set of features (\textit{non-ambiguous}). Due to the low number of ambiguous words in Turkish dataset ($\leq$100), it has been calculated for Finnish only. Similarly, for the derivational morphology analysis in Fig.~\ref{fig:tur}, we have separately calculated scores for sentences containing derived words (\textit{derivational}), and simple sentences without any derivations. Both analyses show that access to gold morphological tags (\textit{oracle}) provided big performance gains on arguments with contextual ambiguity and sentences with derived words. Moderate \textit{IOC} signals that \textit{char} and \textit{char3} learns to imitate the ``beads'' and their ``predictable order'' on the string (in the absence of the aforementioned issues).
    \begin{figure*}
    \centering
      \begin{subfigure}[b]{0.3\textwidth}
        \includegraphics[width=\textwidth]{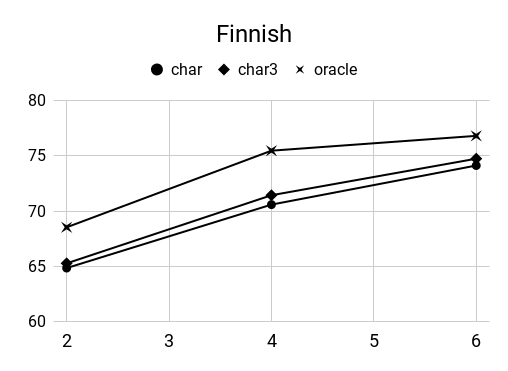}  
    \end{subfigure}
    ~ %add desired spacing between images, e. g. ~, \quad, \qquad, \hfill etc. 
    \begin{subfigure}[b]{0.3\textwidth}
        \includegraphics[width=\textwidth]{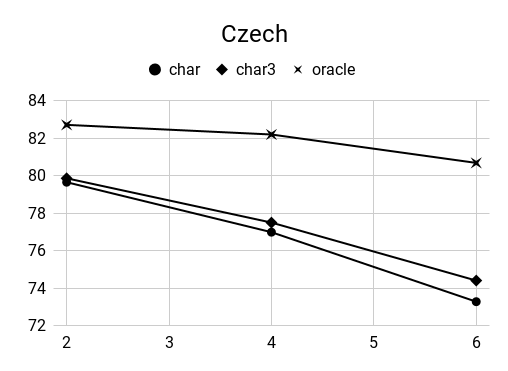} 
    \end{subfigure}
    ~ %add desired spacing between images, e. g. ~, \quad, \qquad, \hfill etc. 
    \begin{subfigure}[b]{0.3\textwidth}
        \includegraphics[width=\textwidth]{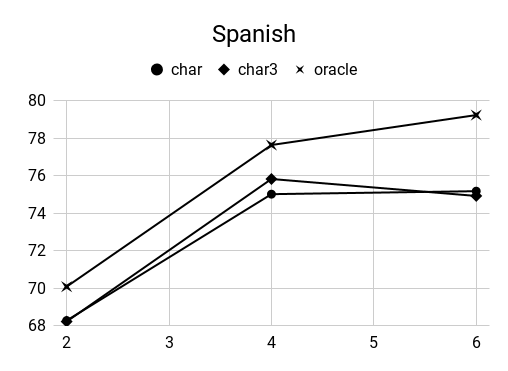}
    \end{subfigure}
    \caption{\textit{x axis}: Number of morphological features; \textit{y axis}: Targeted F1 scores}
   \label{fig:feats}
   \end{figure*}
  \paragraph{Fusional languages} \textit{may} have many morphemes in a word. Spanish and Catalan have relatively low morpheme per word ratio that results with low OOV\% (5.63 and 5.40 for Spanish and Catalan respectively); whereas, German and Czech have OOV\% of 7.93 and 7.98~\cite{conll09}. We observe that \textit{IOW} by character models are well aligned with OOV percentages of the datasets. Unlike agglutinative languages, single morpheme can serve multiple purposes in fusional languages. For instance, ``o''~(e.g., \textit{habl-o}) may signal $1_{st}$ person singular present tense, or $3_{rd}$ person singular past tense. We count the number of surface forms with at least two different features and use their ratio \textit{(\#ambiguous forms/\#total forms)} as a proxy to morphological complexity of the language. The \textit{complexities} are approximated as 22\%, 16\%, 15\% for Czech, Spanish and Catalan respectively; which are aligned with the observed \textit{IOC}s. 
   \begin{table*}
      \scalebox{0.75}
      {
        \begin{tabular}{rccc|ccc|ccc}
         
          & \multicolumn{3}{c}{\textbf{char+char3}} & \multicolumn{3}{c}{\textbf{char+oracle}} & \multicolumn{3}{c}{\textbf{char3+oracle}} \\
         \hline
          & Avg & SG & \textit{IOB\%} & Avg & SG & \textit{IOB\%} & Avg & SG & \textit{IOB\%} \\
         \hline
         \textbf{Czech} & 76.24 & 76.26 & \textbf{\textcolor{higher}{2.03}} & 80.36 & 81.06 & \textit{\textcolor{lower}{0.49}} & 80.57 & 81.10 & \textit{\textcolor{lower}{0.55}}  \\
         \textbf{Finnish} & 70.31 & 70.29 & \textbf{\textcolor{higher}{4.58}} & 72.73 & 72.88 & \textit{\textcolor{lower}{2.42}} & 72.72 & 73.02 & \textit{\textcolor{lower}{2.62}} \\
         \textbf{Turkish} & 59.43 & 59.39 & \textbf{\textcolor{higher}{6.34}} & 61.98 & 62.07 & \textit{\textcolor{lower}{4.53}} & 60.56 & 60.74 & \textit{\textcolor{lower}{2.28}}  \\
         \hline
         \textbf{Spanish} & 70.01 & 70.05 & \textit{\textcolor{lower}{3.16}} & 71.80 & 71.75 & \textbf{\textcolor{higher}{3.47}} & 71.64 & 71.62 & \textbf{\textcolor{higher}{3.24}}  \\
         \textbf{Catalan} & 72.79 & 72.71 & \textit{\textcolor{lower}{2.03}} & 74.80 & 74.82 & \textbf{\textcolor{higher}{2.16}} & 75.15 & 75.18 & \textbf{\textcolor{higher}{2.66}}  \\
         \textbf{German} & 66.84 & 66.97 & \textit{\textcolor{lower}{2.15}} & 71.02 & 71.16 & \textbf{\textcolor{higher}{2.62}} & 71.31 & 71.25 & \textbf{\textcolor{higher}{2.84}}   \\
         \end{tabular}
      }
      \caption{Results of ensembling via averaging (Avg) and stack generalization (SG). \textit{IOB: Improvement Over Best of baseline models}}
    \label{tab:ensem_res}
    \end{table*}
  Since there is no unique morpheme to meaning mapping, generally multiple morphological tags are used to resolve the \textit{morpheme ambiguity}. Therefore there is an indirect relation between the number of morphological tags used and the ambiguity of the word. To demonstrate this phenomena, we calculate targeted F1 scores on arguments with varying number of morphological features. Results using feature bins of [1-2], [3-4] and [5-6] are given in Fig.~\ref{fig:feats}. As the number of features increase, the performance gap between oracle and character models grows dramatically for Czech and Spanish, while it stays almost fixed for Finnish. This finding suggests that high number of morphological tags signal the vagueness/complex cases in fusional languages where character models struggle; and also shows that the complexity can not be directly explained by number of morphological tags for agglutinative languages. German is known for having many compound words and compound lemmas that lead to high OOV\% for lemma; and also is less ambiguous (9\%). Therefore we would expect a lower \textit{IOC}. However, the evaluation set consists only of 550 predicates and 1073 arguments, hence small changes in prediction lead to dramatic percentage changes.  
 
  \subsection{Similarity between models}
    %We have hypothesized that character based models learn to imitate the morphological process up to a certain degree. 
    One way to infer similarity is to measure \textit{diversity}. Consider a set of baseline models that are not diverse, i.e., making similar errors with similar inputs. In such a case, combination of these models would not be able to overcome the biases of the learners, hence the combination would not achieve a better result. In order to test if character and morphological models are \textit{similar}, we combine them and measure the performance of the ensemble. Suppose that a prediction $p_{i}$ is generated for each token by a model $m_i$, $i \in n$, then the final prediction is calculated from these predictions by:
    \begin{gather}
      p_{final} =  f(p_0, p_1,..,p_n|\phi)
    \end{gather} 
    where $f$ is the combining function with parameter $\phi$. The simplest global approach is \textit{averaging (AVG)}, where $f$ is simply the mean function and $p_i$s are the log probabilities. Mean function combines model outputs linearly, therefore ignores the nonlinear relation between base models/units. In order to exploit nonlinear connections, we learn the parameters $\phi$ of $f$ via a simple linear layer followed by sigmoid activation. In other words, we train a new model that learns how to best combine the predictions from subword models. This ensemble technique is generally referred to as \textit{stacking} or \textit{stacked generalization (SG)}. %and is illustrated in Fig.~\ref{sfig:late}. 
    ~\footnote{To train the SG model, we have used one linear layer with 64 hidden units followed by sigmoid nonlinear activation. Weights are orthogonally initialized and optimized via adam algorithm with a learning rate of 0.02 for 25 epochs.} 

    Although not guaranteed, diverse models can be achieved by altering the input representation, the learning algorithm, training data or the hyperparameters. To ensure that the only factor contributing to the diversity of the learners is the input representation, all parameters, training data and model settings are left unchanged. 

    Our results are given in Table~\ref{tab:ensem_res}. \textit{IOB} shows the improvement over the best of the baseline models in the ensemble. Averaging and stacking methods gave similar results, meaning that there is no immediate nonlinear relations between units. We observe two language clusters: (1) Czech and agglutinative languages (2) Spanish, Catalan, German and English. The common property of that separate clusters are (1) high OOV\% and (2) relatively low OOV\%. Amongst the first set, we observe that the improvement gained by character-morphology ensembles is higher~(shown with green) than ensembles between characters and character trigrams~(shown with red), whereas the opposite is true for the second set of languages. It can be interpreted as character level models being more similar to the morphology level models for the first cluster, i.e., languages with high OOV\%, and characters and morphology being more diverse for the second cluster.
         
\section{Limitations and Strengths}
      To expand our understanding and reveal the limitations and strengths of the models, we analyze their ability to handle long range dependencies, their relation with training data and model size; and measure their performances on out of domain data. 
        \begin{figure*}
        \centering
          \begin{subfigure}[b]{0.3\textwidth}
            \includegraphics[width=\textwidth]{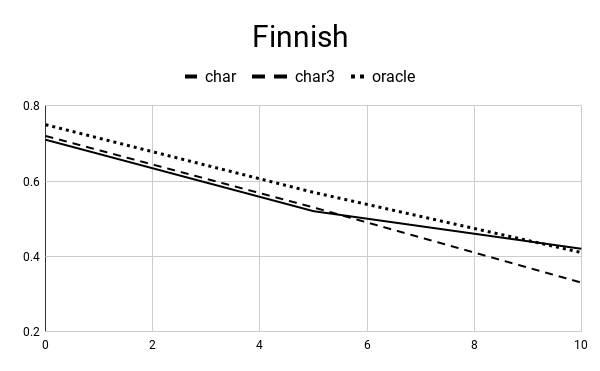}  
        \end{subfigure}
        ~ %add desired spacing between images, e. g. ~, \quad, \qquad, \hfill etc. 
        \begin{subfigure}[b]{0.3\textwidth}
            \includegraphics[width=\textwidth]{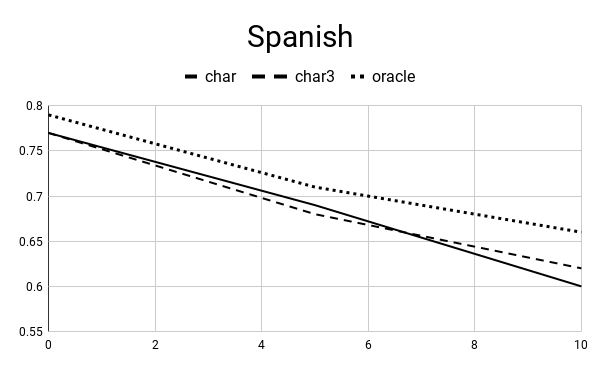} 
        \end{subfigure}
        ~ %add desired spacing between images, e. g. ~, \quad, \qquad, \hfill etc. 
        \begin{subfigure}[b]{0.3\textwidth}
            \includegraphics[width=\textwidth]{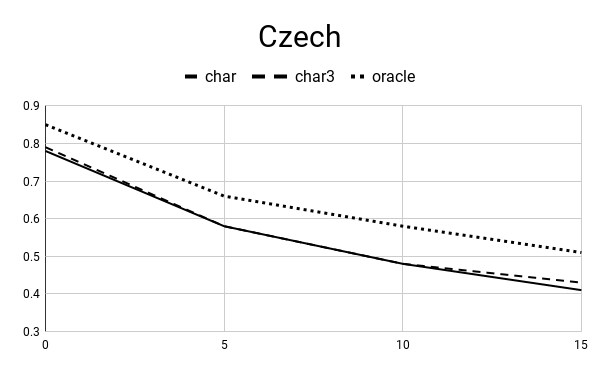}
        \end{subfigure}
        \caption{\textit{X axis}:~Distance between the predicate and the argument, \textit{Y axis}: F1 scores on argument labels}
       \label{fig:dtp_analysis}
       \end{figure*}

      \subsection{Long Range Dependencies}
          Long range dependency is considered as an important linguistic issue that is hard to solve. Therefore the ability to handle it is a strong performance indicator. To gain insights on this issue, we measure how models perform as the distance between the predicate and the argument increases. The unit of measure is number of tokens between the two; and argument is defined as the head of the argument phrase in accordance with dependency-based SRL task. For that purpose, we created bins of [0-4], [5-9], [10-14] and [15-19] distances. Then, we have calculate F1 scores for arguments in each bin. Due to low number of predicate-argument pairs in buckets, we could not analyze German and Turkish; and also the bin [15-19] is only used for Czech. Our results are shown in Fig.~\ref{fig:dtp_analysis}. We observe that either \textit{char} or \textit{char3} closely follows the \textit{oracle} for all languages. The gap between the two does not increase with the distance, suggesting that the performance gap is not related to long range dependencies. In other words, both characters and the oracle handle long range dependencies equally well.

      \subsection{Training Data Size}
\begin{figure*}
            \centering
              \begin{subfigure}[b]{0.32\textwidth}
                \includegraphics[width=\textwidth]{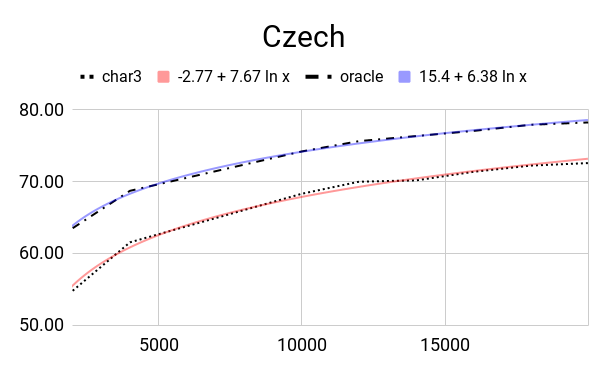}  
            \end{subfigure}
            ~ %add desired spacing between images, e. g. ~, \quad, \qquad, \hfill etc. 
            \begin{subfigure}[b]{0.32\textwidth}
                \includegraphics[width=\textwidth]{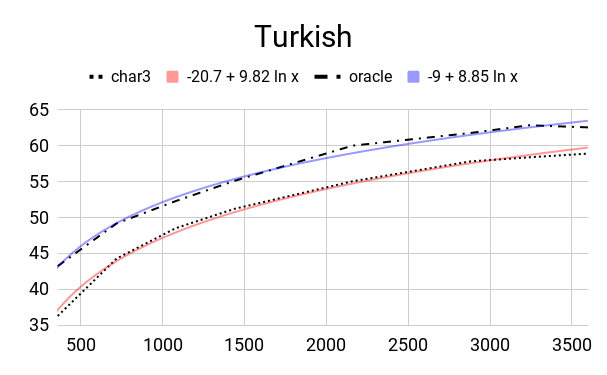} 
            \end{subfigure}
            ~ %add desired spacing between images, e. g. ~, \quad, \qquad, \hfill etc. 
            \begin{subfigure}[b]{0.32\textwidth}
                \includegraphics[width=\textwidth]{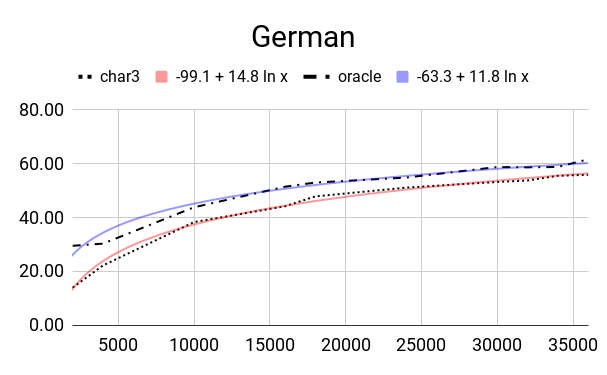}
            \end{subfigure}
            \caption{Performance of units w.r.t training data size. \textit{X axis}: Number of sentences, \textit{Y axis}: F1 score}
           \label{fig:ds_analysis}
          \end{figure*}

          We analyzed how \textit{char3} and \textit{oracle} models perform with respect to the training data size. For that purpose, we trained them on chunks of increasing size and evaluate on the provided test split. We used units of 2000 sentences for German and Czech; and 400 for Turkish. Results are shown in Fig.~\ref{fig:ds_analysis}. Apparently as the data size increases, the performances of both models logarithmically increase - with a varying speed. To speak in statistical terms, we fit a logarithmic curve to the observed F1 scores (shown with transparent lines) and check the \textit{x} coefficients, where \textit{x} refers to the number of sentences. This coefficient can be considered as an approximation to the speed of growth with data size. We observe that the coefficient is higher for \textit{char3} than \textit{oracle} for all languages. It can be interpreted as: in the presence of more training data, \textit{char3} may surpass the \textit{oracle}; i.e., \textit{char3} relies on data more than the \textit{oracle}. 

      \subsection{Out-of-Domain (OOD) Data}
          As part of the CoNLL09 shared task~\cite{conll09}, out of domain test sets are provided for three languages: Czech, German and English. We test our models trained on regular training dataset on these OOD data. The results are given in Table~\ref{tab:ood_res}. Here, we clearly see that the best model has shifted from oracle to character based models. The dramatic drop in German oracle model is due to the high lemma OOV rate which is a consequence of keeping compounds as a single lemma. Czech oracle model performs reasonably however is unable to beat the generalization power of the \textit{char3} model. Furthermore, the scores of the character models in Table~\ref{tab:ood_res} are higher than the best OOD scores reported in the shared task~\cite{conll09}; even though our main results on evaluation set are not (except for Czech). This shows that character-level models have increased robustness to out-of-domain data due to their ability to learn regularities among data.
          \begin{table}
            \scalebox{0.60}
            {
            \begin{tabular}{|l|c|c|c|c|c|c|c|c|}
             \hline
               & \textbf{word}  &  \textbf{char} & \textit{IOW\%} &  \textbf{char3} & \textit{IOW\%} &  \textbf{oracle} & \textit{IOW\%} & \textit{IOC\%} \\
               \hline
               \textbf{CZE} & 69.97 & 72.98 & \textit{4.30} & \textbf{73.24} & \textit{4.67} & 72.28 & \textit{3.30} & \textit{-1.31} \\
               \hline             
               \textbf{DEU} & 51.50 & \textbf{57.05} & \textit{10.78} &  55.75 & \textit{8.24} & 38.51 & \textit{-25.24} & \textit{-45.17} \\
               \hline
               \textbf{ENG} & 66.47 & 68.83 & \textit{0.70} & \textbf{70.22} & \textit{0.23} & - & - & - \\   
               \hline
             \end{tabular}
            }
            \caption{F1 scores on out of domain data. Best scores are shown with \textbf{bold}.}
         \label{tab:ood_res}
         \end{table}

      \subsection{Model Size}
         Throughout this paper, our aim was to gain insights on how models perform on different languages rather than scoring the highest F1. For this reason, we used a model that can be considered small when compared to recent neural SRL models and avoided parameter search. However, we wonder how the models behave when given a larger network. To answer this question, we trained \textit{char3} and \textit{oracle} models with more layers for two fusional languages (Spanish, Catalan), and two agglutinative languages (Finnish, Turkish). The results given in Table~\ref{tab:modelsize_res} clearly shows that model complexity provides relatively more benefit to morphological models. This indicates that morphological signals help to extract more complex linguistic features that have semantic clues. 
            \begin{table}
            \scalebox{0.75}
            {
              \begin{tabular}{rrcc|cc}
               & & \multicolumn{2}{c}{\textbf{char3}} & \multicolumn{2}{c}{\textbf{oracle}} \\
                 & & F1 & \textit{I (\%)} & F1 & \textit{I (\%)} \\
                 \hline 
                 \multirow{2}{*}{\textbf{Finnish}} & $\ell=1$ & 67.78 & & 71.15 & \\ 
                                                   & $\ell=2$ & 67.62 & \textit{-0.2} & 75.71 & \textit{6.4}\\ 
                 \hline
                 \multirow{2}{*}{\textbf{Turkish}} & $\ell=1$ & 56.60 & & 59.38 & \\ 
                                                   & $\ell=2$ & 56.93 & \textit{0.5} & 61.02 & \textit{2.7} \\ 
         
                 \hline
                 \multirow{2}{*}{\textbf{Spanish}} & $\ell=1$ & 68.43 & & 69.39 & \\ 
                                                   & $\ell=2$ & 69.30 & \textit{1.3} & 71.56 & \textit{3.1}\\ 
                 \hline

                 \multirow{2}{*}{\textbf{Catalan}} & $\ell=1$ & 71.34 & & 73.24 & \\ 
                                                   & $\ell=2$ & 71.71 & \textit{0.5} & 74.84 & \textit{2.2}\\ 
                 \hline
               \end{tabular}
            }
            \caption{Effect of layer size on model performances. \textit{I}: Improvement over model with one layer.}
          \label{tab:modelsize_res}
          \end{table}

      \subsection{Predicted Morphological Tags}
\begin{figure}
              \centering  
              \includegraphics[scale=0.25]{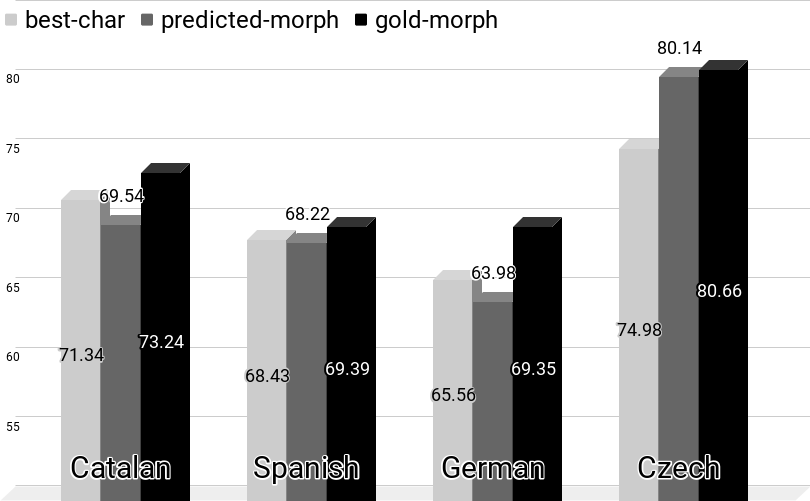}
              \caption{F1 scores for \textit{best-char} (best of the CLMs) and model with predicted (\textit{predicted-morph}) and gold morphological tags (\textit{gold-morph}).}
              \label{fig:pred_morph}
          \end{figure}

          Although models with access to gold morphological tags achieve better F1 scores than character models, they can be less useful a in real-life scenario since they require gold tags at test time. To predict the performance of morphology-level models in such a scenario, we train the same models with the same parameters with predicted morphological features. Predicted tags were only available for German, Spanish, Catalan and Czech. Our results given in Fig.~\ref{fig:pred_morph}, show that (except for Czech), predicted morphological tags are not as useful as characters alone. 

\section{Conclusion}
  Character-level neural models are becoming the \textit{defacto} standard for NLP problems due to their accessibility and ability to handle unseen data. In this work, we investigated how they compare to models with access to gold morphological analysis, on a sentence-level semantic task. We evaluated their quality on \textit{semantic role labeling} in a number of agglutinative and fusional languages. Our results lead to the following conclusions:
  \begin{itemize}
  \item For in-domain data, character-level models cannot yet match the performance of morphology-level models. However, they still provide considerable advantages over whole-word models, 
  \item Their shortcomings depend on the morphology type. For agglutinative languages, their performance is limited on data with rich \textit{derivational morphology} and high \textit{contextual ambiguity} (morphological disambiguation); and for fusional languages, they struggle on tokens with high number of morphological tags,    
  \item Similarity between character and morphology-level models is higher than the similarity within character-level (char and char-trigram) models on languages with high OOV\%; and vice versa,
  \item Their ability to handle long-range dependencies is very similar to morphology-level models,
  \item They rely relatively more on training data size. Therefore, given more training data their performance will improve faster than morphology-level models,
  \item They perform \textit{substantially} well on out of domain data, surpassing all morphology-level models. However, relatively less improvement is expected when model complexity is increased,
  \item They generally perform better than models that only have access to predicted/silver morphological tags.
  \end{itemize}

\section{Acknowledgements}
Gözde Gül Şahin was a PhD student at Istanbul Technical University and a visiting research student at University of Edinburgh during this study. She was funded by Tübitak (The Scientific and Technological Research Council of Turkey) 2214-A scholarship during her visit to University of Edinburgh. She was granted access to CoNLL-09 Semantic Role Labeling Shared Task data by Linguistic Data Consortium (LDC). This work was supported by ERC H2020 Advanced Fellowship GA 742137 SEMANTAX and a Google Faculty award to Mark Steedman. We would like to thank Adam Lopez for fruitful discussions, guidance and support during the first author's visit.

%\bibliography{acl2018}

\bibliography{acl2018}
\bibliographystyle{acl_natbib}

\end{document}